\documentclass[lettersize,journal]{IEEEtran}
\usepackage{amsmath,amsfonts}
\usepackage{algorithmic}
\usepackage{algorithm}
\usepackage{array}
\usepackage[caption=false,font=normalsize,labelfont=sf,textfont=sf]{subfig}
\usepackage{textcomp}
\usepackage{stfloats}
\usepackage{url}
\usepackage{verbatim}
\usepackage{graphicx}
\usepackage{cite}
\usepackage{makecell}
\usepackage{color}
\usepackage{booktabs}
\usepackage{multirow}
\newcommand{\z}{{\rm\bf z}}                   
\newcommand{\w}{{\rm\bf w}}                   
\newcommand{\X}{{\rm\bf X}}                   

\begin{document}
\title{StarNet: Style-Aware 3D Point Cloud Generation}

\author{Yunfan Zhang$^*$, Hao Wang$^*$, Guosheng Lin, Vun Chan Hua Nicholas, Zhiqi Shen, Chunyan Miao
\thanks{Yunfan Zhang, Hao Wang, Guosheng Lin, Vun Chan Hua Nicholas, Zhiqi Shen, and Chunyan Miao are with School of Computer Science and Engineering, Nanyang Technological University.
E-mail: \{yunfan001, hao005, gslin, aschvun, zqshen, ascymiao\}@ntu.edu.sg.}
 \thanks{Corresponding authors: Guosheng Lin, Vun Chan Hua Nicholasr and Shiqi Shen.}
}

\markboth{Journal of \LaTeX\ Class Files,~Vol.~14, No.~8, August~2021}%
{Shell \MakeLowercase{\textit{et al.}}: A Sample Article Using IEEEtran.cls for IEEE Journals}


\maketitle
\def\thefootnote{*}\footnotetext{These authors contributed equally to this work}\def\thefootnote{\arabic{footnote}}

\begin{abstract}
This paper investigates an open research task of reconstructing and generating 3D point clouds. Most existing works of 3D generative models directly take the Gaussian \textit{prior} as input for the decoder to generate 3D point clouds, which fail to learn disentangled latent codes, leading noisy interpolated results. Most of the GAN-based models fail to discriminate the local geometries, resulting in the point clouds generated not evenly distributed at the object surface, hence degrading the point cloud generation quality. Moreover, prevailing methods adopt computation-intensive frameworks, such as flow-based models and Markov chains, which take plenty of time and resources in the training phase.
To resolve these limitations, this paper proposes a unified style-aware network architecture combining both point-wise distance loss and adversarial loss, StarNet which is able to reconstruct and generate high-fidelity and even 3D point clouds using a mapping network that can effectively disentangle the Gaussian \textit{prior} from input’s high-level attributes in the mapped latent space to generate realistic interpolated objects.  
Experimental results demonstrate that our framework achieves comparable state-of-the-art performance on various metrics in the point cloud reconstruction and generation tasks, but is more lightweight in model size, requires much fewer parameters and less time for model training.
\end{abstract}

\begin{IEEEkeywords}
Point Cloud, 3D, Generation, Style-aware
\end{IEEEkeywords}

\section{Introduction}
\IEEEPARstart{P}OINT cloud is a simple representation of 3D objects using spatial XYZ coordinates as $N\times3$ matrices and/or surface norms of $N\times6$. It is one of the most important 3D object representations and now widely used in various applications, such as robotics \cite{wu2016fast, jiayao2022real}, remote sensing \cite{zhang2017advances, nguyen2019robust}, and autonomous driving \cite{hu2021graph, yu2022rate}, to represent objects in the real world. In these applications, the obtained point clouds from the sensors are usually incomplete and sparse. Therefore, further processing such as point cloud reconstruction is necessary and vital for machine perception, which demonstrates the significance of reconstructing the high-resolution and high-fidelity point clouds. The ability to generate diversified and high-fidelity point cloud objects could enlarge the scarce high-quality point cloud dataset and can be used for data augmentation purpose as well. Moreover, in creative design works, e.g. interior design \cite{li2018application}, the capability to generate various point cloud objects gives more freedom to designers when they draft their ideas.

\begin{figure}[t]
\centering
\includegraphics[width=\columnwidth]{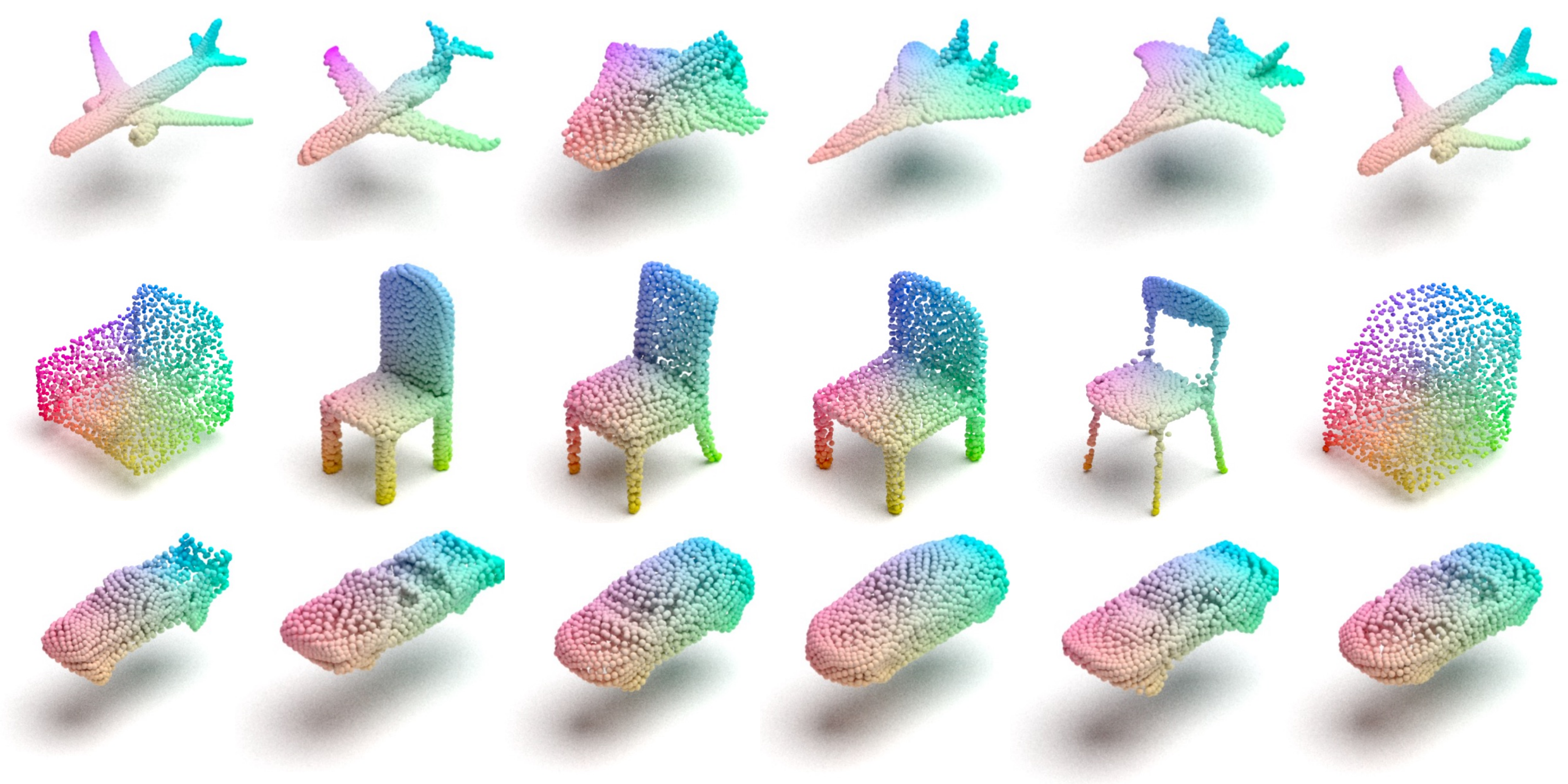}
\caption{The visualization of generated 3D airplanes, chairs and cars whose prior are sampled from Gaussian \textit{prior}. Each object consists of 2048 points and each point is visualized as a small sphere.}
\label{fig:gen_v0}
\end{figure}

The unordered nature of 3D point clouds makes it challenging to process them directly. Point cloud voxelization \cite{wu2016learning, choy20163d} are mainly used in some earlier works. However, voxel-based methods take much time and consume a lot of memory during the training phase. Meanwhile, quantization errors are also introduced during voxelization. The recent works \cite{achlioptas2018rgan,yang2019pointflow,cai2020shapegf,luo2021diffusion} propose to apply generative models directly on the point clouds instead of using the voxel-based methods. To be specific, ShapeGF \cite{cai2020shapegf} learns the gradient field with point clouds disturbed by stochastic noise and reconstructs point clouds through Langevin dynamics \cite{welling2011bayesian} moving along these gradients. \cite{luo2021diffusion} proposes a diffusion-based model using a series of Markov chains. 

The aforementioned 3D generative models follow similar structures to the conventional 2D image counterparts, which sample Gaussian \textit{prior} as the input to generate the 3D objects. However, conventional models fail to learn disentangled latent codes and cannot produce realistic enough images. StyleGAN \cite{karras2019stylebased} is proposed to address these limitations. Here we extend the idea of StyleGAN to 3D point cloud generation. Technically, we introduce a mapping network to map the Gaussian \textit{prior} to another latent space such that features in the mapped latent space are disentangled. This practice enables us to control the semantics of the input latent codes, which means we can better manipulate the generated outputs through interpolation. Moreover, the aforementioned works are either large in their model size or incorporating complicated operations, making them not easy to train.

In this paper, we propose an efficient and effective framework: Style-Aware 3D Point Cloud Reconstruction and Generation framework, StarNet in short. It incorporates a two-stage design: auto-encoder consisting of a point cloud encoder and a novel style-aware decoder for the point cloud reconstruction and a generative network consisting of a mapping network and the pre-trained style-aware decoder in the auto-encoder for the point cloud generation. The point cloud encoder is PointNet-based \cite{qi2017pointnet} to learn the point cloud representation and encode the point clouds into vectors. Inspired by StyleGAN \cite{karras2019stylebased}, our novel style-aware decoder consists of the affine transformation from latent codes to \textit{styles} codes, Adaptive Instance Normalization (AdaIN) \cite{huang2017arbitrary} to transfer \textit{styles} codes to object contents, and the Squeeze-and-Excitation (SE) \cite{hu2019squeezeandexcitation} layers to control information flow by weighting each channel adaptively. Earth Mover's Distance (EMD) and Chamfer Distance (CD) are widely adopted to evaluate point cloud reconstruction. We propose to optimize an objective function combining both EMD and CD. For the point cloud generation, we incorporate the adversarial training method \cite{goodfellow2014gan} to train the above-mentioned point cloud generator together with a PointNet-based point cloud discriminator. Some of the generated point clouds are illustrated in Figure \ref{fig:gen_v0}. Our proposed framework does not incorporate computational-complex modules or operation like other frameworks but it still outperforms previous works.

The following are our main contributions: 
\begin{itemize}
\setlength{\itemsep}{0pt}
    \item We propose a novel unified style-aware framework to reconstruct and generate visual-pleasing and high-fidelity point clouds.
    \item Incorporating both point-wise reconstruction loss and adversarial loss, our generative model can generate diversified point cloud objects whose points are evenly distributed, and closer to the distribution of the point clouds in dataset.
    \item Our framework is lightweight, efficient, and effective with smaller model size, it requires less training resource compared with existing works.
    \item Our model achieves state-of-the-art results on various metrics in point cloud reconstruction and generation.
\end{itemize}

\begin{figure*}[t]
\begin{center}
\includegraphics[width=\textwidth]{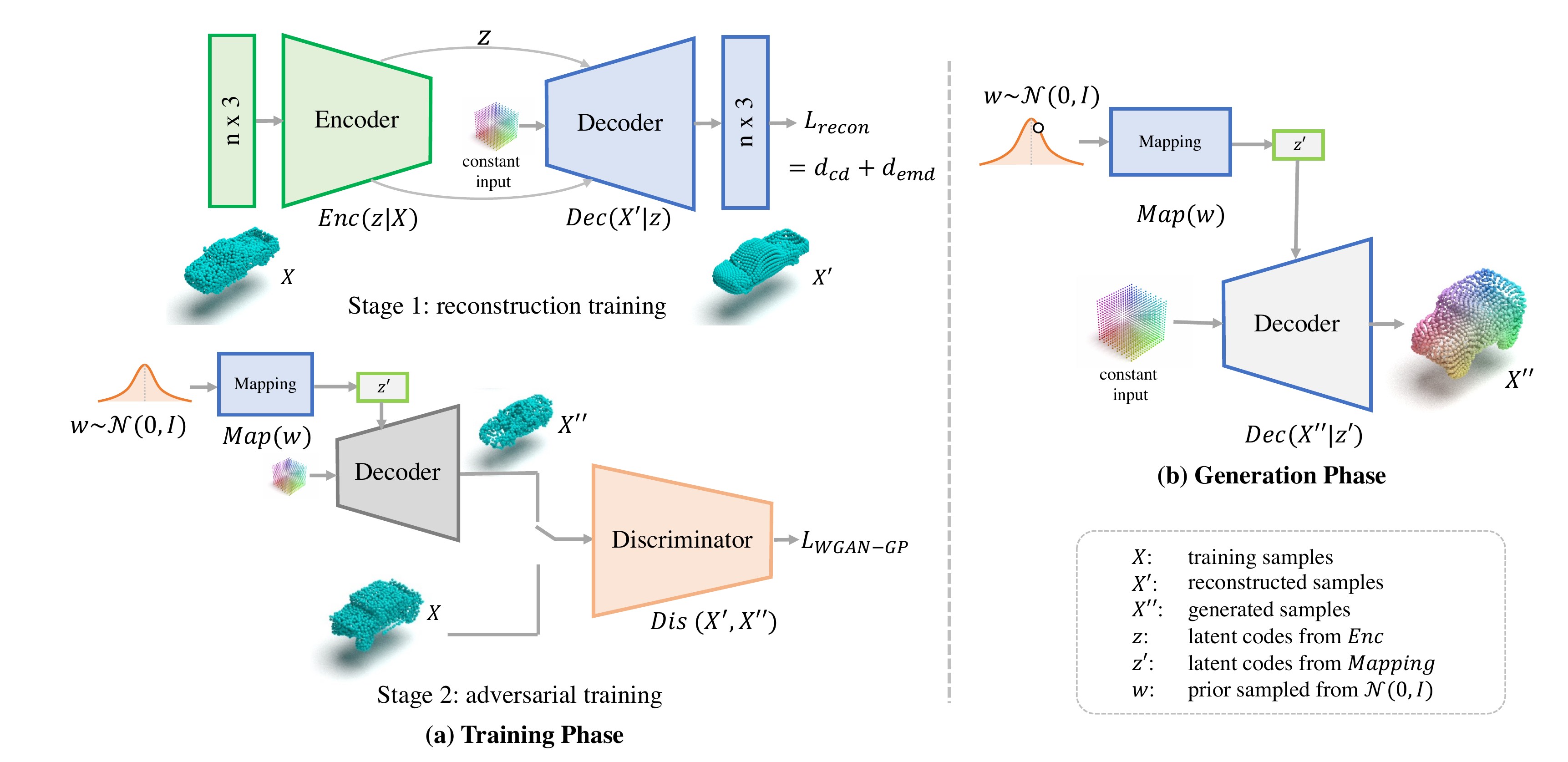}
\end{center}
\caption{The illustration of the StarNet framework. (a) The two-stage training process: point cloud auto-encoding and generation. In Stage 1, we adopt point cloud encoder $Enc(\z|\X)$ to encode $X$ to $z$. Then, we feed $\z$ and a constant input into the style-aware decoder $Dec(\X'|\z)$, where we use the cubic like point clouds as the constant input to be the starting point. We propose the combined CD and EMD as our training objective $L_{recon}$. In Stage 2, we input $\w$ to the mapping network $Map(\w)$ to generate $\z'$ and decode $\z'$ to generate object $\X''$ through the pre-trained style-aware decoder $Dec(\X''|\z')$ with the help of the point cloud discriminator $Dis(\X, \X'')$. We train Stage 2 using WGAN-GP $L_{WGAN-GP}$. (b) The generation process: $\w$ is sampled from Gaussian \textit{prior} and fed into the trained $Map(\w)$ and $Dec(\X''|\z')$ from (a) to generate point cloud objects.}
\label{fig:framework}
\end{figure*}

\section{Related Works}
\label{sec:related}
3D shape generation has been drawing more and more interest in the community. There are various tasks related such as 3D shape reconstruction from single-view image, 3D shape completion from partial 3D input and 3D shape transformations. In this work, we focus on the 3D point cloud generation. Specifically, it learns the distribution from the training dataset and generates novel and diverse point clouds that are not in the training dataset but still are high-fidelity and visual pleasing.

\subsection{Point cloud generative models}
The permutation-invariance nature of point clouds makes it hard to process them directly. Voxel-based methods process the dataset through voxelization of point clouds. Voxelization provides a way to transform the unordered data structure to the grid-like data format, which can be further processed using 3DCNN like operations. Voxel-based generative models have been proven successful \cite{girdhar2016learning,wu2016learning, maimaitimin2017stacked,hinks2013point,meng2019vv}, whereas, these methods are computational and memory expensive as one more dimension introduced compared to 2D counterparts. Thus, these methods usually cannot be scaled to high resolution to generation high fidelity point clouds and they suffer from information lost during data format conversion.

PointNet \cite{qi2017pointnet} provides a way to process the point clouds directly without voxelization. By projecting the low dimensional data into high dimensional space and using symmetrical functions, PointNet preserves the permutation invariant nature of point clouds. In light of this, various generative models deal with point clouds directly are invented such as autoregressive models,  generative adversarial models and auto-encoding models.

Autoregressive-based generative models learns the joint 3D point distribution in a point cloud. PointGrow \cite{sun2020pointgrow} is an autoregressive-baed model to estimate the point distributions. During generation phase, the point clouds are growing iteratively based on the previous points generated. Whereas the iterative generation nature of autoregressive models makes the generation stage slow and they cannot scale well to higher point cloud resolution.

GAN based generative models learn the data distribution in an adversarial way. There are various GAN models to generate point clouds with the help of PointNet-based discriminator. \cite{li2018point} is the first to propose a GAN-based model: rGAN and l-GAN to generate the point clouds from a Gaussian \textit{prior}. l-GAN only operate the bottleneck latent codes learnt and extracted from the encoder. \cite{valsesia2018gcngan} is another GAN variant incorporating the Graph-Convolution Network. \cite{shu20193d} is a tree-structured GAN model performs graph convolution within a tree. \cite{hui2020pdgn} proposes a progressively generative GAN model that generates dense point clouds in a coarse-to-fine manner. These models not only incur complex modules in their generators, but also generate results suffering from the uneven points distribution or even clustering as the discriminator only incorporates \textit{max} pooling which only extract global features. 

Other generative models like flow-based models and diffusion-based models are emerging recently. They learn the distribution of points in a point cloud object with the help of the invertible parametric transformations. PointFlow \cite{yang2019pointflow} proposes a flow-based generative model using a VAE framework through a series of continuous normalizing flow, which heavily relies on the expensive Ordinary Differential Equation (ODE) solver. \cite{cai2020shapegf} proposes to learn the gradient of the  point clouds disturbed by different noise levels and generate point clouds along the learnt gradient. Most recently, \cite{luo2021diffusion} proposes a diffusion model inspired by the diffusion process in non-equilibrium thermodynamics modeled by the Markov Chain. It takes significant time to train these models and it is also very slow to generate point cloud objects in a multi-step manner. Moreover, the invertibility constraint in flow-based approach limits the representation learning capability and the series transformations tends to introduce an averaging effect based on training data distribution, as a result, the generated point clouds tend to noisy as depicted in the later part.

\subsection{StyleGAN}
StyleGAN \cite{karras2019stylebased} achieves tremendous success in image generation. One of the key points of StyleGAN is introducing a mapping network to map Gaussian \textit{prior} to the latent space. As such features in latent space are disentangled. In contrast to traditional generative models which take Gaussian \textit{prior} as the first stage input and generate images randomly without controllable features, StyleGAN takes a constant input into the first layer and the latent codes to each layer through an affine transformation with AdaIN \cite{huang2017arbitrary} to transfer the \textit{styles} information. This architecture can control image generation in a multi-scale manner. As such, this model is capable not only of generating impressively photorealistic high-quality images, but also offers control over the \textit{styles} of the generated image at different levels of detail by changing the \textit{styles} vectors. In StyleGAN, Wasserstein GAN with Gradient Policy (WGAN-GP) \cite{gulrajani2017improved} is adopted to stabilize GAN training procedure by matching the Wasserstein distance and penalizing the gradient norm to avoid gradient vanishing problem.

\section{Method}
\label{sec:method}
In this section, we formulate our framework into two-stage design: auto-encoding for point cloud reconstruction and GAN for point cloud generation. The overall framework is depicted in Figure \ref{fig:framework}. 

The training pipeline is presented in Figure \ref{fig:framework}(a), which can be summarized in a two-stage training procedure:
\begin{itemize}
   \item \textbf{Stage 1.} We train the auto-encoder to learn the point cloud representation in latent space $\z$ as well as to reconstruct the point clouds conditioned on these codes using the set distance loss.
   \item \textbf{Stage 2.} We use WGAN-GP training strategy to train the point cloud generator with the style-aware decoder well-trained in stage 1 and a PointNet-based point cloud discriminator.
\end{itemize}

The generation pipeline is presented in Figure \ref{fig:framework}(b). Gaussian \textit{prior} $\w$ is sampled initially before feeding to the mapping network to generate the latent codse $\z'$. Afterward, the style-aware decoder decodes the codes $\z'$ to synthesis the corresponding point cloud objects $\X''$.

\subsection{Point Cloud Auto-encoder}
\paragraph{Encoder} 
The encoder in the auto-encoder takes the unprocessed point clouds as the raw input and encode dataset $\X$ of $N\times2048\times3$ to latent space $\z$ as $Enc(\z|\X)$ such that $\z\sim\mathbb{R}^{128}$, where $N$ is number of training objects in the dataset. We use both \textit {max} and \textit {avg} pooling as the symmetric functions in the encoder to achieve the best representation. 

\begin{figure}[t]
\begin{center}
\includegraphics[width=0.8\columnwidth]{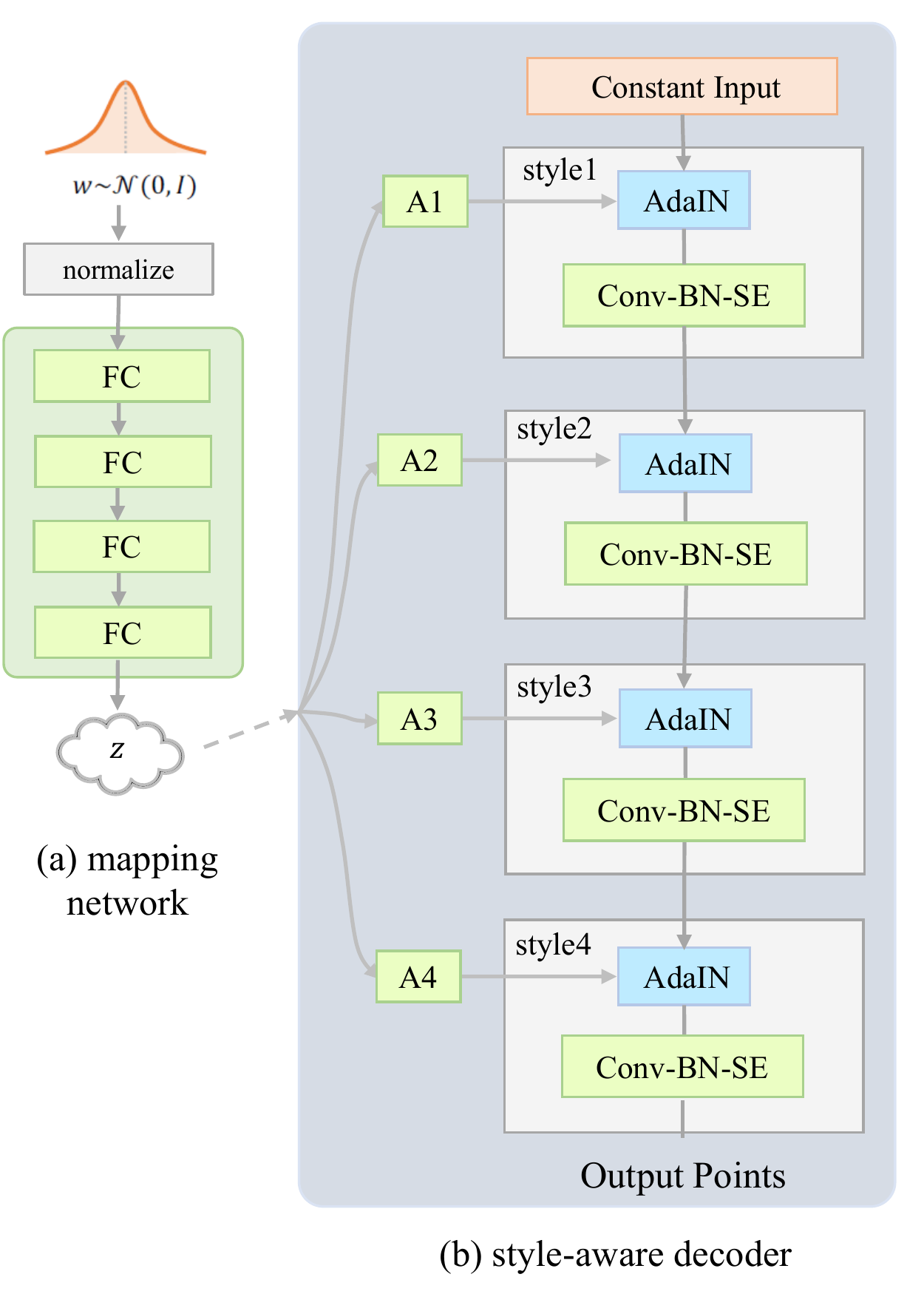}
\end{center}
\caption{The illustration of the StarNet generator consisting of the mapping network and the style-aware decoder.}
\label{fig:decoder}
\end{figure}

\paragraph{Decoder}
The novel style-aware decoder is to decode the latent codes $\z$ to reconstruct the point cloud objects $\X'$ of $N\times2048\times3$ as $Dec(\X'|\z)$. The decoder consists of an affine transformation to transfer the latent codes $\z$ to the \textit{style} codes feeding into each layer, a constant input of $2048\times3$ uniformly sampled from unit cube as the starting point of generation, and the core networks aggregating the \textit{styles} information to object contents. Detail implementation of the decoder is depicted in Figure \ref{fig:decoder}(b). Specifically, in each convolution layer, AdaIN \cite{karras2019stylebased} is to transfer \textit{styles} codes to object contents, followed by a 1D convolution operation, batch-normalization (BN) \cite{ioffe2015batch} layer, and the SE layer. All activation functions are \textit{LeakyReLU} with slope of 0.2 except for the final layer of \textit{tanh} to generate point cloud objects.

\paragraph {Adaptive Instance Normalization}
The latent codes $\z\sim\mathbb{R}^{128}$ pass through the affine transformation to generate the \textit{styles} codes, which are different in dimension in each layer. The affine transformation is implemented using an MLP neural network to generate the scaling $\lambda$ and biasing $\beta$. Following the style-transfer technique, \cite{huang2017arbitrary} proposes the AdaIN defined as Eq: \ref{eq:adain}. 

\begin{equation}
\label{eq:adain}
    AdaIN(x_i, y)=y_{s, i}\frac{x_i - \mu(x_i)}{\sigma(x_i)}+y_{b,i}
\end{equation}

Each channel of the feature maps $x_i$ is firstly normalized to zero mean and unit variance. Then, the normalized feature maps are demodulated by the scaling and biasing. These demodulated feature maps modify the relative importance of features for current layers and they do not depend on the previous layers because of the normalization. The constant input acts as the starting content, and the \textit{styles} codes guide the object generation in each layer conditioned on the latent codes $\z$. Feature maps can be controlled in different scales and more independently \cite{karras2019stylebased}. 

\paragraph{Squeeze-and-Excitation Layer}
Due to the unorderness of point clouds, dynamic edge convolution \cite{wang2019dynamic} is widely adopted to aggregate feature information. The computational cost is quite high due to k-Nearest Neighbours (kNN) involved. Instead, we incorporate Squeeze-and-Excitation (SE) layer \cite{hu2019squeezeandexcitation, xie2021stylebased} in our decoder. The SE layer uses the content-aware mechanism to weight each channel separately to realize the control information flow. A global descriptor of each channel is squeezed into a single numeric value by \textit{max} pooling each channel of the feature maps. Then, it is fed through a two fully-connected layers, which outputs a same size vector. After going through a \textit{sigmod} layer, these values are now used as weights on the original features maps, scaling each channel based on its importance.

\subsection{Point Cloud GAN}

\paragraph{Generator}
The generator, as depicted in Figure \ref{fig:decoder}, consists of the mapping network and the pre-trained style-aware decoder. The mapping network is a MLP neural network to map gaussion \textit{prior} to the latent codes. It is used to provide an isolation between the Gaussian \textit{prior} and the latent codes, thus, the features in the latent space are well disentangled, leading to the noise-free latent codes interpolation. As depicted in Figure \ref{fig:decoder}(a), the mapping network consists of four fully-connected layers with batch normalization in between, following the similar structure in StyleGAN \cite{karras2019stylebased}. The pre-trained style-aware decoder is adopted to synthesis the point cloud objects conditioned on the latent codes generated from the mapping network. The latent codes is passed through the affine transformations to define multi-scale \textit{styles} code, which are integrated in the generator via AdaIN. The use of this \textit{styles} vector gives multi-level control over the style of the generated point cloud objects.

\paragraph{Discriminator}
The point cloud discriminator adopts the similar PointNet-based structure as the above-mentioned point cloud encoder. In the discriminator, only \textit{max} pooling is adopted as the symmetric function and output is a 1D vector to critic whether the point cloud objects are from dataset or generated by the generator. We use \textit{LeakyReLU} with slope of 0.2 and batch normalization in every layer, similar to \cite{shu2019treegan, hui2020pdgn}. We update the discriminator 5 times before update the generator once.

\subsection{Training Objectives}
\paragraph{Auto-encoding} ShapeGF \cite{cai2020shapegf} uses L2 loss to train a field gradient. PointFlow \cite{yang2019pointflow} and DPM \cite{luo2021diffusion} model use statistical likelihood as the reconstruction loss without working on point distance loss. Auto-encoding is usually qualified by the point distance metrics Chamfer Distance (CD) and Earth Mover's Distance (EMD), thus, it is in favour to optimize them directly. CD is to measure the average matching distance to the nearest neighbour, whereas EMD, known as Wasserstein metric is to measure the minimum cost to transport from one distribution to another. The implementation of EMD in \cite{achlioptas2018rgan, yang2019pointflow} uses a greedy approximation $O(n^2)$ of the Hungarian algorithm. Recently, \cite{liu2020morphing} implements an auction-based EMD with $O(n)$ memory and $O(n^2k)$ time complexity, where $k$ is the number of iterations. This novel implementation makes it possible to train the model faster and more efficient using EMD as the reconstruction loss. We propose to combine CD and EMD as the compound reconstruction loss depicted in Eq: \ref{eq:loss} to achieve the best reconstruction performance. 

\begin{equation}
\label{eq:loss}
    L_{recon} = d_{cd}(X, Y) + d_{emd}(X, Y)
\end{equation}

\paragraph{Generation} Conventional GAN suffers from mode collapse and gradient vanishing problem \cite{arjovsky2017wasserstein}. WGAN \cite{arjovsky2017wasserstein} proposes to use Wasserstein distance as the cost function as it has a smoother gradient everywhere. WGAN-GP \cite{gulrajani2017improved} further stabilizes the training with a novel clipping methods by penalizing the norm of gradient of the discriminator with respect to its input, keeping norm of weights gradient around 1. The training object of WGAN-GP is 

\begin{equation}\label{eq:losswgan}
    L = \underset{\tilde{x} \sim \mathbb{P}_g}{\mathbb{E}}[D(\tilde{x})-\underset{x \sim \mathbb{P}_r}{\mathbb{E}}[D(x)]]+ \lambda\underset{\hat{x} \sim \mathbb{P}_{\hat{x}}}{\mathbb{E}}[(||\bigtriangledown_{\hat{x}} D(\hat{x})||_2-1)^2]
\end{equation}

\section{Experiments}

In this section, we evaluate our framework in both point cloud auto-encoding and generation on ShapeNet \cite{shapenet2015} dataset.

\subsection{Datasets} 

ShapeNet is a large-scale dataset of 3D shapes consisting of 55 categories with 51,127 unique models, such as airplane, chair, desk etc. We follow 85-15 \cite{luo2021diffusion} split for training and testing in each category. Each object is sampled 2048 points from the shape and is re-centered to the origin and normalized into the $[-1, 1]^3$ cube following the same way as \cite{cai2020shapegf}.

\subsection{Evaluation Metrics}

\paragraph{Point Cloud Auto-encoding} Following prior work \cite{achlioptas2018rgan}, CD and EMD are adopted to evaluate the reconstruction performance. Compared to CD, reconstructed point clouds with lower EMD tend to be more faithful with higher visual quality \cite{achlioptas2018rgan}.

\paragraph{Point Cloud Generation} To quantify the point cloud generation performance, we adopt the evaluation metrics: Jensen-Shannon Divergence (JSD),  Minimum Matching Distance (MMD) and Coverage (COV) introduced by \cite{achlioptas2018rgan}. JSD measures a marginal point distributions between reference set and generated set. MMD is an average distance between two point cloud sets to measure the fidelity between them. COV is a percentage metric that measures the how well the points in the reference set is matched in the generated set. It can be used to detect mode collapse but not the quality of the generated point clouds. Both CD and EMD are used to calculate MMD and COV.

\subsection{Point Cloud Auto-encoding}

\begin{figure}[t]
\begin{center}
\includegraphics[width=\columnwidth]{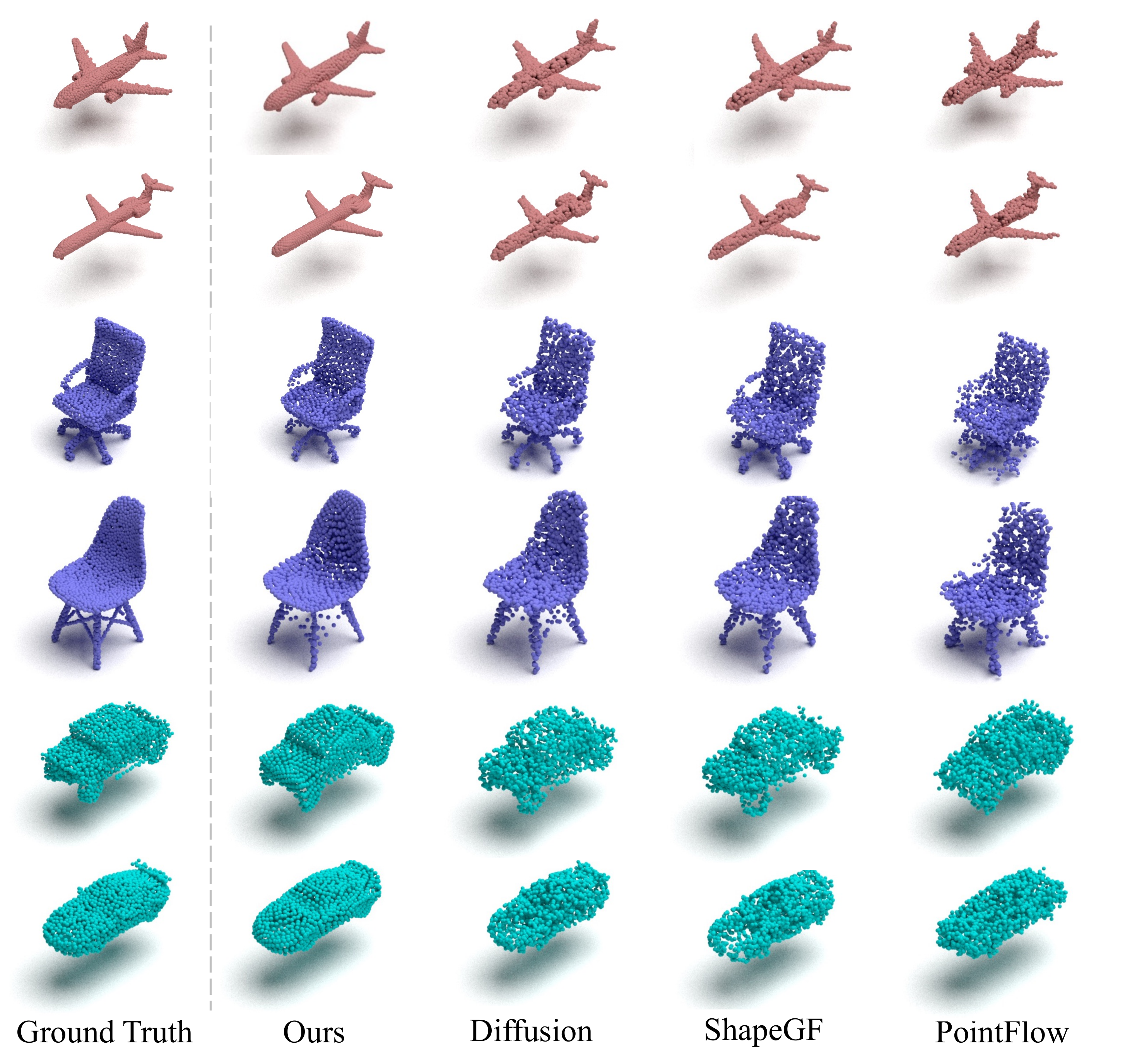}
\end{center}
\caption{Point cloud reconstruction comparison, the first column shows the the ground truth inputs, the rest are the reconstructed results from different models.}
\label{fig:recon_comp}
\end{figure}

\begin{table*}[t]
\caption{The comparison of \textbf{point cloud auto-encoding} performance. CD is multiplied by $10^4$ and EMD is multiplied by $10^2$.}
\begin{center}
\resizebox{\textwidth}{!}{
\begin{tabular}{l c|ccccc c|c}
\toprule
Dataset                   & Metric ($\downarrow$) & AtlasNet (S1) & AltasNet (P25) & PointFlow & ShapeGF & DPM & Ours & Oracle           \\ 
\midrule
\multirow{2}{*}{Airplane} & CD   & 2.000  & \bf{1.795} & 2.420& 2.102   & 2.118 & 2.038 & 1.016 \\
                          & EMD   & 4.311 & 4.336 & 3.311& 3.508   & 2.876& \bf{2.532} & 2.141 \\
\midrule
\multirow{2}{*}{Car}      & CD  & 6.906   & 6.503 & 5.828& 5.468   & 5.493& \bf{4.854} & 3.917 \\
                          & EMD  & 5.617  & 5.408 & 4.390& 4.489   & 3.937& \bf{3.374} & 3.246 \\
\midrule
\multirow{2}{*}{Chair}    & CD   & 5.479  & \bf{4.980} & 6.795& 5.146   & 5.677& 5.595 & 3.221 \\
                          & EMD  & 5.550  & 5.282 & 5.008& 4.784   & 4.153& \bf{4.015} & 3.281 \\
\midrule
\multirow{2}{*}{ShapeNet} & CD   & 5.873  & 5.420 & 7.550& 5.725   & 5.252& \bf{4.959} & 3.074 \\
                          & EMD  & 5.457  & 5.599 & 5.172& 5.049   & 3.783& \bf{3.571} & 3.112 \\
\bottomrule
\end{tabular}}
\end{center}
\label{table:ae}
\end{table*}

\begin{figure}[t]
\begin{center}
\includegraphics[width=\columnwidth]{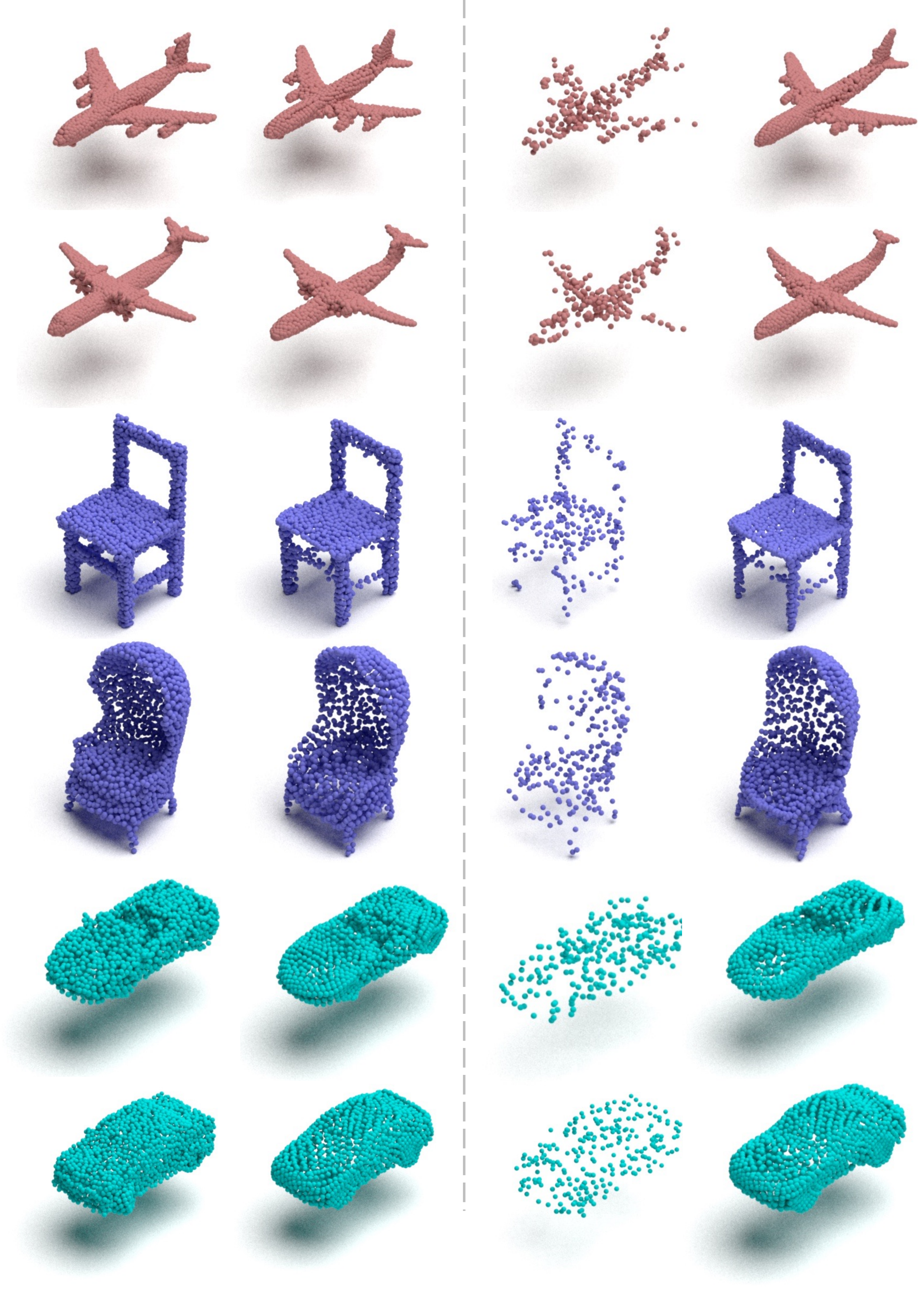}
\end{center}
\caption{Point cloud reconstruction results. The left column: we input 2048 points and output 2048 points. The right column: we input 256 points sampled from the same shape and output 2048 points.}
\label{fig:recon}
\end{figure}

\paragraph{Training Details} We train and evaluate our auto-encoder on four categories: airplane, car, chair and the whole ShapeNet. We use Adam \cite{kingma2017adam} optimizer with initial learning rate 0.001 and betas of (0.9, 0.99). We adopt the step learning rate scheduler to decay the learning rate after 400 epochs with ratio 0.1. We use batch size 128 and train each category 500 epochs. The reconstructed point clouds are de-normalized using their scaling and biasing recorded during data pre-processing. The same evaluation code\footnote{https://github.com/stevenygd/PointFlow} is adopted as \cite{yang2019pointflow} for apple-to-apple comparison. The comparison of reconstruction results with the recent works are tabulated in Table \ref{table:ae}. The lower limit bound "oracle" is obtained by computing the distance between two different samples from the same object. Some visualizations comparison are depicted in Figure \ref{fig:recon_comp}.

EMD is more reliable to distinguish the visual quality of the point clouds. The results with smaller EMD show the points are distributed more evenly \cite{liu2020morphing}, showing higher fidelity. From table \ref{table:ae}, our method consistently outperforms all other methods in both CD and EMD metrics in all four categories by quite a big of margin, narrowing the gap to the lower limit bound "oracle". Moreover, Figure \ref{fig:recon_comp} demonstrates that our model reconstructs the most clean, evenly distributed and visual-pleasing point cloud objects. Overall, our auto-encoder achieves state-of-the-art performance in point cloud reconstruction.

\begin{figure*}[t]
\begin{center}
\includegraphics[width=\textwidth]{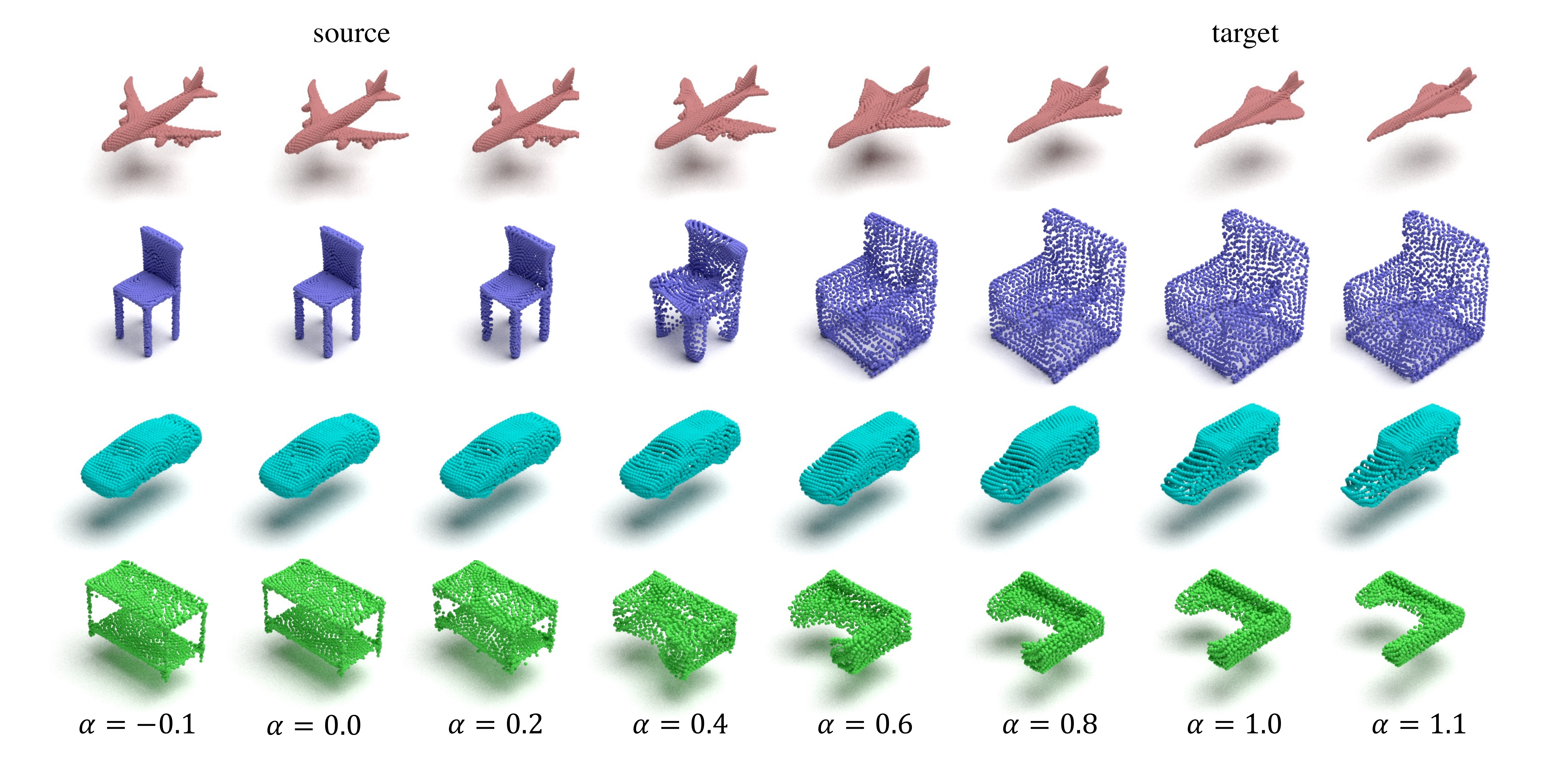}
\end{center}
\caption{The visualization of feature space $\z$ interpolation and extrapolation. $\alpha=0.0$ and $\alpha=1.0$ are the source and target objects. The shapes in between are generated by interpolating the two shapes in space $\z$ and outside ones are extrapolated.}
\label{fig:lerp}
\end{figure*}

\paragraph{Point Cloud Up-sampling} Following \cite{cai2020shapegf}, we conduct another reconstruction experiment taking 256 point clouds as the auto-encoder input to check the up-sampling performance. The reconstruction results are illustrated in Figure \ref{fig:recon_comp}. Even though there are some details missing in the second car and second chair examples, the overall results are visually faithful and comparable to those of 2048 inputs.

\paragraph{Interpolation and Extrapolation in Latent Space} We also conduct the latent codes interpolation and extrapolation following \cite{yang2019pointflow, luo2021diffusion}. The $\alpha$ is the interpolation and extrapolation ratio. $\alpha=0.0$ corresponds to the source latent code and $\alpha=1.0$ to the target. They are obtained by $Enc(\z|\X)$ taking the ground-truth point clouds as inputs. The intermediate codes are linearly interpolated in step size of 0.2.  Extrapolation are done on both side. The style-aware decoder $Dec(\X''|\z)$ synthesises the corresponding 3D points clouds conditioned on these codes. Examples from four categories are depicted in Figure \ref{fig:lerp}. Every object in the plot is noise-free and vivid in detail. Taking airplane as an example, the source is a four-engine passenger airplane and the target is a fighter jet. The intermediate results show these engines are gradually disappearing, the spoiler is shrinking, and the wings are growing from source to target. The similar trends continue in the extrapolations. Similar behaviour appears in chair, car and the whole ShapeNet dataset. Our auto-encoder learns good point cloud representation and reconstructs visually faithful results.

\subsection{Unsupervised representation learning}
We conduct the similar experiment to evaluate the representation learning of our auto-encoder following the experimental setting \cite{luo2021diffusion, yang2019pointflow}. Firstly, our auto-encoder is trained on the whole ShapeNet dataset. Secondly, the latent codes are extracted from the well-trained encoder. Finally, the latent codes are then used to train an SVM classifier on ModelNet10 or ModelNet40 \cite{modelnet}. During training, the point clouds are normalized to zero-mean and unit variance globally, and point clouds are augmented by random rotation along gravity axis. The ModelNet classification results on the test split are depicted in the table \ref{table:classify}. We use their numbers reported in the corresponding papers for comparison. The performance of our auto-encoder is comparable to recent generative models.

\begin{table}[t]
\caption{Comparison of representation learning in linear SVM classification accuracy.}
\label{table:classify}
\begin{center}
\begin{tabular}{l|cc}
\toprule
\makecell[c]{Model} & ModelNet10 & ModelNet40 \\
\midrule
AtlasNet \cite{groueix2018atlasnet} 
    & 91.9 & 86.6 \\
l-GAN (CD) \cite{achlioptas2018rgan}
    & 95.4 & 84.5 \\
l-GAN (EMD) \cite{achlioptas2018rgan}
    & 95.4 & 84.0 \\
PointFlow \cite{yang2019pointflow}
    & 93.7 & 86.8 \\
ShapeGF \cite{cai2020shapegf}
    & 90.2 & 84.6 \\
DPM \cite{luo2021diffusion}
    & 94.2 & 87.6 \\
\midrule
Ours
    & 93.3 & 86.5 \\
\bottomrule
\end{tabular}

\end{center}
\end{table}

\subsection{Point Cloud Generation}

\begin{figure}[t]
\centering
\includegraphics[width=\columnwidth]{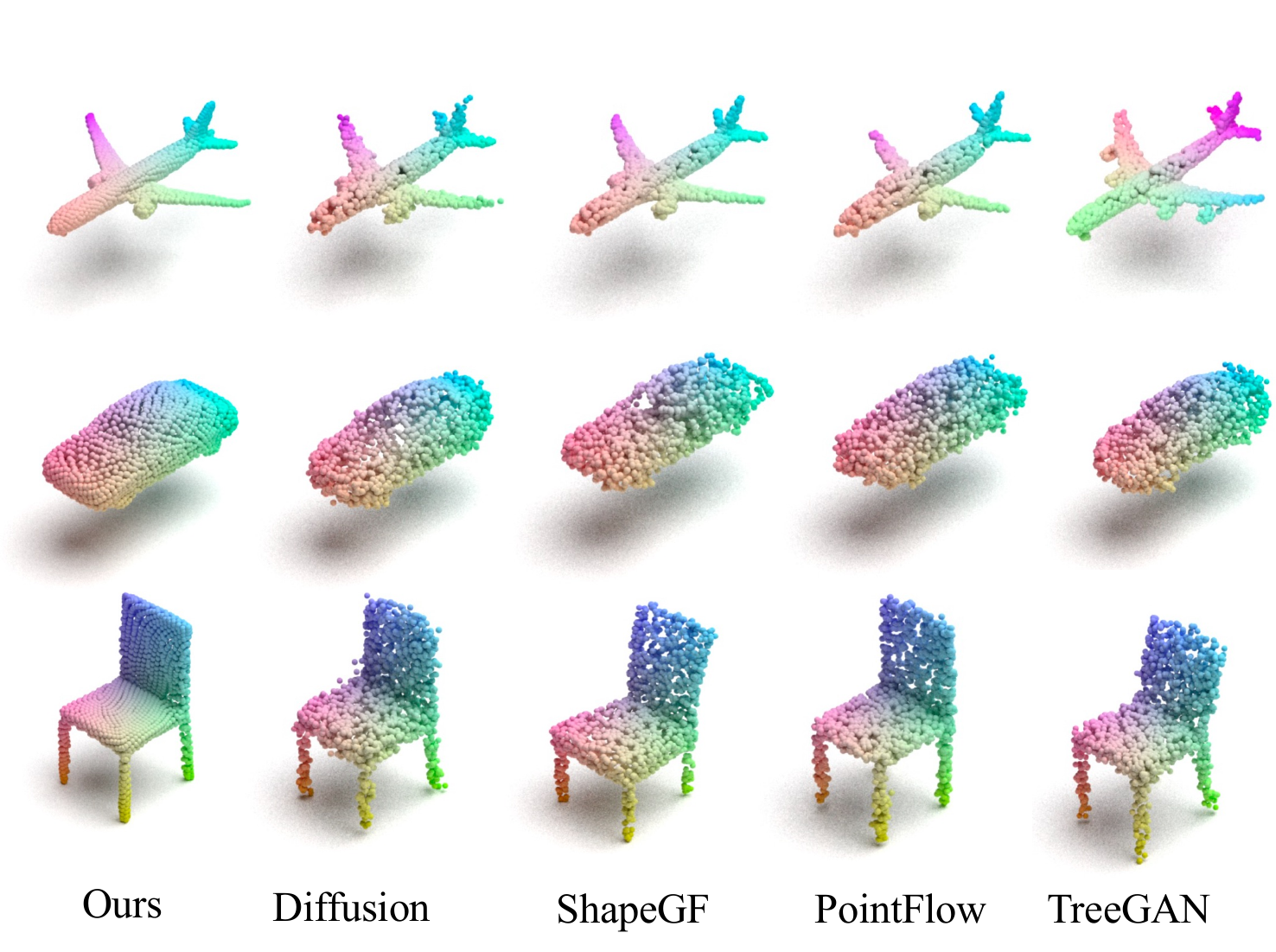}
\caption{The generated 3D airplanes, chairs and cars from different models.}
\label{fig:gen_comp}
\end{figure}

\begin{figure*}[t]
\centering
\includegraphics[width=\textwidth]{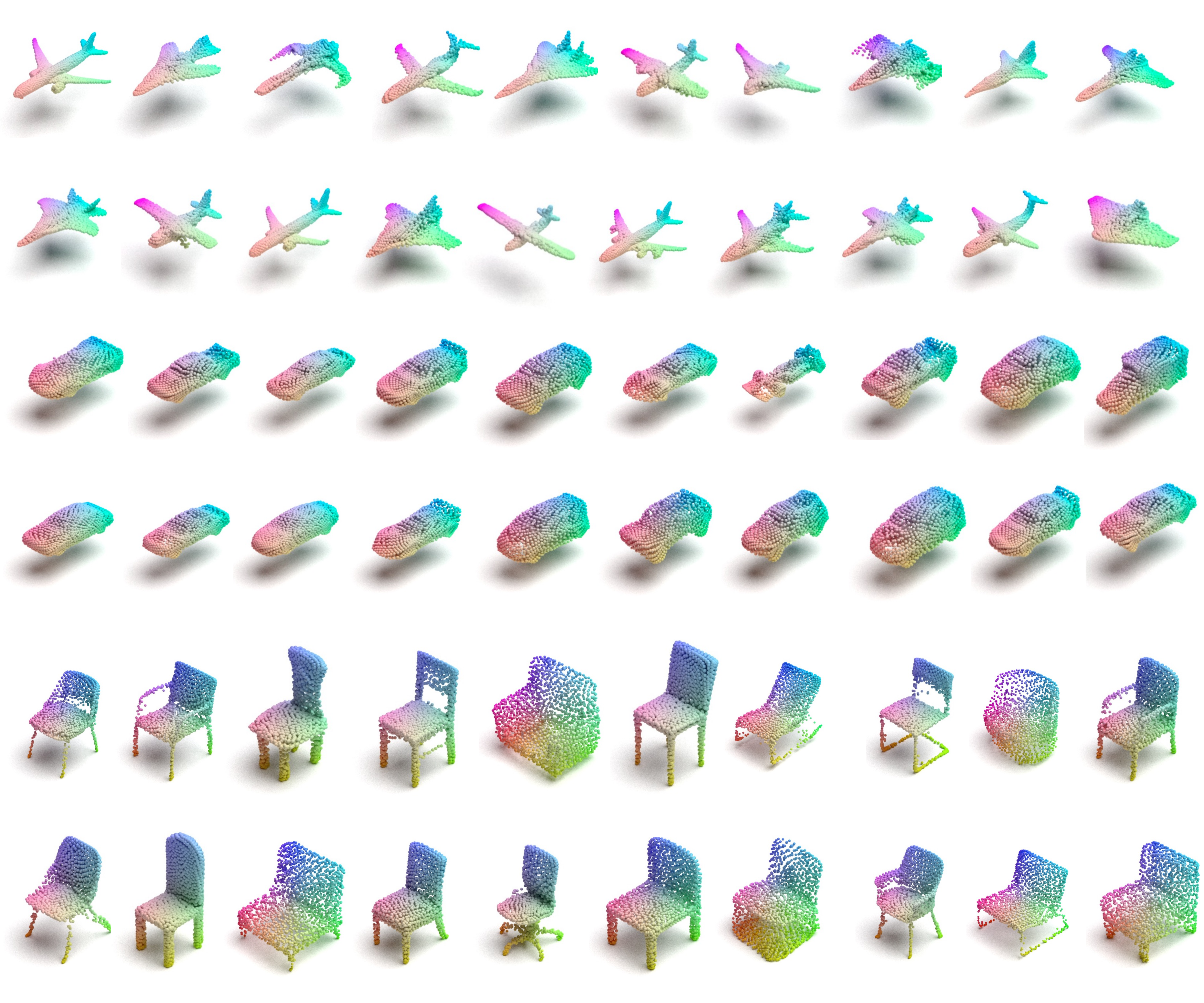}
\caption{The generated 3D airplanes, chairs and cars whose priors are sampled from Gaussian \textit{prior}. The generated results demonstrate diversified and visually plausible shapes in each category.}
\label{fig:gen}
\end{figure*}

\begin{table*}[t]
\caption{The comparison of \textbf{point cloud generation} performance for recent models. $\uparrow$: the higher the better, $\downarrow$: the lower the better. The bold numbers are the best results. MMD-CD scores are multiplied by $10^4$; MMD-EMD scores are multiplied by $10^2$; JSDs are multiplied by $10^2$. Parameters and Time are number of trainable parameters in (Mb) and approximate training time in (hours). PointFlow has fewer parameters than ours, but it incorporates non-trainable ODE solver and takes almost 10 times longer training time.}
\begin{center}
\resizebox{0.85\textwidth}{!}{

\begin{tabular}{ll|c|cc|cc|cc}

\toprule
 &  & JSD ($\downarrow$) & \multicolumn{2}{c|}{MMD ($\downarrow$)} & \multicolumn{2}{c|}{COV (\%, $\uparrow$)} & \multicolumn{2}{c}{1-NNA (\%, $\downarrow$)} \\ \cmidrule{3-9}
Shape & Model & - & CD & EMD & CD & EMD & CD & EMD \\
\midrule

\multirow{7}{*}{Airplane} 
 & PC-GAN \cite{achlioptas2018rgan} & 6.188 & 3.819 & 1.810 & 42.17 & 13.84 & 77.59 & 98.52  \\
 & GCN-GAN \cite{valsesia2018gcngan} & 6.669 & 4.713 & 1.650 & 39.04 & 18.62 & 89.13 & 98.60  \\
 & TreeGAN \cite{shu2019treegan} & 15.646 & 4.323 & 1.953 & 39.37 &  8.40 & 83.86 & 99.67  \\
 & PointFlow \cite{yang2019pointflow} & 1.536 & 3.688 & 1.090 & 44.98 & 44.65 & 66.39 &  \textbf {69.36} \\
 & ShapeGF \cite{cai2020shapegf} &  1.314 &  3.727 &  1.061 & 46.79 &  44.08 &  \textbf {62.69} &  71.91 \\ 
 & DPM \cite{luo2021diffusion} & 1.067 & \textbf {3.276} &  1.061 &  \textbf {47.78} & 44.32 &  64.83 & 75.12 \\ 
\cmidrule{2-9}
 & \textbf {Ours} & \textbf{0.992} & \textcolor{blue}{\textbf{3.300}} & \textbf{0.906} &	\textcolor{blue}{\textbf{47.62}} & \textbf{44.81} & 70.92 & 74.95 \\
\midrule

\multirow{7}{*}{Chair}
 & PC-GAN \cite{achlioptas2018rgan} & 6.649     & 13.436 & 3.104 & 46.23 & 22.14 & 69.67 & 100.00 \\
 & GCN-GAN \cite{valsesia2018gcngan} & 21.708   & 15.354 & 2.213 & 39.84 & 35.09 & 77.86 & 95.80 \\
 & TreeGAN \cite{shu2019treegan}  & 13.282     & 14.936 & 3.613 & 38.02 &  6.77 & 74.92 & 100.00 \\
 & PointFlow \cite{yang2019pointflow}  & 12.474    & 13.631 & 1.856 & 41.86 & 43.38 & 66.13 & 68.40 \\ 
 & ShapeGF \cite{cai2020shapegf}  & 5.996    & 13.175 & 1.785 &  48.53 &  46.71 & \textbf {56.17} & \textbf {62.69} \\ 
 & DPM \cite{luo2021diffusion} &  7.797 & 12.276 & 1.784 & 48.23 &  47.52 & 60.11 & 69.06 \\ 
\cmidrule{2-9}
 & \textbf {Ours}  & \textbf {4.324}     & 	\textbf {12.075}   & 	\textbf {1.590}   & 	\textbf {48.84}   & 	\textbf {47.52}   & \textcolor{blue}{\textbf{60.11}}   & \textcolor{blue}{\textbf{64.21}} \\
\bottomrule
\end{tabular}}
\end{center}
\label{table:gen}
\end{table*}

\paragraph{Training Details} In generation, we adopt the pre-trained style-aware decoder in auto-encoder and update the parameters in the mapping network $Map(\w)$ and the point cloud discriminator using WGAN-GP strategy. Following the prior work \cite{gulrajani2017improved}, Adam optimizer is adopted using a fixed learning rate 0.0001 and betas (0.5, 0.9). The weight of gradient policy is 10. We use batch size 64 and stop training after 500 epochs.

To make fair comparison to the state-of-the-art work \cite{luo2021diffusion}, we use the same evaluation codes\footnote{https://github.com/luost26/diffusion-point-cloud} as \cite{luo2021diffusion}. Specifically, we generate the same number of objects as in the test set and compare these two sets of point clouds. Both of these two sets are normalized. The evaluation results are reported in Table \ref{table:gen}. In both airplane and chair category, our model achieves achieves state-of-the-art performance. From the visualization illustrated in Figure \ref{fig:gen}, Our model can synthesis the point clouds more uniformly distributed and diversified. The generated point cloud with similar shapes from different models are depicted in Figure \ref{fig:gen_comp} as well. It shows that our results generate the most evenly distributed and realistic, and cleanest point clouds.

We also tabulated the number of trainable parameters for recent models in both the full framework and the generative part in Table \ref{table:gen}. In StarNet, the generative part consists of the mapping network and the style-aware decoder with parameters of 0.5M and 1.3M each, thus, 1.35M in total, making it the second smallest model in the table. It is slightly larger than PointFlow \cite{yang2019pointflow}. However, PointFlow incorporates the complicated ODE solver slowing the training procedure.

Despite of the two-stage design, our model is efficient and takes less time to train, including both auto-encoding and generation. Taking airplane as an example, the reported training time of \cite{cai2020shapegf} and \cite{yang2019pointflow} are around one day and two days respectively. Whereas it only takes less than 4 hours to train our two-stage framework, which are more than 6 and 12 times faster. Based on this, our whole model is smaller in size and more efficient to train, making it a potential framework in various applications.

\subsection{Ablation Study}

\begin{table}[t]
\begin{center}
\caption{Settings for \textbf{auto-encoding} ablation study: \textbf{A.} decoder using MLP only; \textbf{B.} the style-aware decoder without SE layers; \textbf{C.} SE layers are changed to dynamic edge convolution with 10 neighbors support; \textbf{D.} constant input in a 2D surface format; \textbf{E.} loss function using CD only; and \textbf{F.} loss function using EMD only. Parameters and Time denote the trainable parameters count in (Mb) and the training time for 500 epochs in hour, respectively. CD is multiplied by $10^4$ and EMD is multiplied by $10^2$.}
\resizebox{\columnwidth}{!}{
\begin{tabular}{l|ccccc}
\toprule
Setting             & Parameters	        & 	Memory      & 	Time        & 	CD	     & 	EMD  \\
\midrule
\textbf{A}. MLP 	    & 	6.91        & 	\bf 4.99	& \bf 1.17		&   2.356	 &  2.926 \\
\textbf{B}. w/o SE	    & 	\bf{1.13}	& 	11.53	    & 	2.02	    & 	2.244	 & 	2.515 \\
\textbf{C}. w/ k-NN     & 	28.66	    & 	14.02	    & 	7.52	    & 	2.039	 & 	2.607 \\
\textbf{D}. Surface   	& 	1.29	    & 	13.18	    & 	2.01	    & 	2.110	 & 	2.781 \\
\textbf{E}. w/ CD 	    & 	1.29	    & 	13.28	    & 	2.01	    & 	2.038	 & 	3.881 \\
\textbf{F}. w/ EMD 	    & 	1.29	    & 	13.28	    & 	1.86	    & 	2.112	 & 	2.533 \\

\midrule
\bf Full model  & 	\textcolor{blue}{\textbf{1.29}}	 & 	13.28	 & 	2.02	 & 	\bf 2.038	 & \bf	2.532 \\
\bottomrule
\end{tabular}
}
\end{center}
\label{table:ablation}
\end{table}

We conduct ablation experiments in the point cloud reconstruction as depicted in Table \ref{table:ablation}. Each row in the table corresponds to one setting. All experiments are conducted on Tesla V100 16G single GPU platform. The batch size is 128. The time is recorded after 500 epochs. We observe that all the ablation settings cause the drop of reconstruction metrics in CD and EMD comparing to StarNet shown at the bottom. \textbf{A} is considered as our baseline incorporating an MLP-based decoder. \textbf{B} is StarNet without SE layers, so it is slightly smaller whereas its performance drops. \textbf{C} is the largest model with kNN included and it takes more than 3 times training time. Its EMD is far away from that of StarNet. \textbf{D} uses uniform surface as input, resulting in worse result. As StarNet is trained by optimizing the combination of CD and EMD, it achieves the best performance while maintaining comparable model size and training time.

\begin{table}[h]
\caption{Settings for \textbf{generation} ablation study. In \textbf{A}, the generator is trained from scratch without pre-training the style-aware decoder. In \textbf{B}, the generator consists of style-aware decoder only without mapping network.}
\begin{center}
\resizebox{0.96\columnwidth}{!}{
\begin{tabular}{l|c|cc|cc}
\toprule
\multirow{2}{*}{Settings} & \multirow{2}{*}{JSD} & \multicolumn{2}{c|}{MMD} & \multicolumn{2}{c}{COV} \\ \cmidrule{3-6}
 & & CD & EMD & CD & EMD\\
\midrule
\textbf{1}. no pre-train   & 1.621 & 4.239	 & 1.847 &	40.24	& 37.53 \\
\textbf{2}. no mapping  & 27.39  & 144.2 & 21.72 & 21.25 & 20.59 \\
\midrule
\bf StarNet           &    \bf{0.957} & \bf{3.306} & \bf 0.944 & \bf{48.43} & \bf{47.44} \\
\bottomrule
\end{tabular}
}
\end{center}
\label{table:ablation_gen}
\end{table}

We also conduct the ablation experiment in point cloud generation by training the GAN starting from scratch without pre-training the style-aware decoder, we find these points in the objects generated tend to gather into clusters, results in drastic performance drop in all generative evaluation metrics. StarNet incorporates both point-wise reconstruction loss and adversarial loss, achieving best performance in quality and diversity.

\subsection{Limitations}
\paragraph{Discriminator} Most of the pure GAN-based models adopt the similar PointNet-based discriminator structure. In the discriminator adopts \textit{max} pooling as the symmetric function, which is in favor of the global context. Thus, the local context features cannot be discriminated properly. It can be visualized in Figure \ref{fig:gen_comp} that the GAN-based results demonstrate the clustering of points, while still maintain good object shapes. In one of our ablation study settings, when our GAN is trained from scratch, the results also show the similar phenomenon, leading to the degradation of the generation performance. Our framework solves this issue by combining the point-wise distance loss and adversarial loss, however, it is still in the way of a two-stage design. By discriminating both global context and local context in the adversarial way, the model could be trained in an end-to-end manner.
\paragraph{Evaluation Metrics for Generation} Each metric in the generation evaluation has its limitation. The JSD measures the similarity of two distributions whereas it favors the averaging effect that only considers the whole set distributions but not the distribution of individual shapes. MMD tries to find every point in the reference set the nearest neighbour in the generated set. COV could be used to detect mode collapse. However, both MMD and CD could not explicitly determine the quality of the generated point clouds.

\section{Conclusion}
Our proposed 3D-StarNet is an effective and efficient point cloud reconstruction and generation framework that can reconstruct high quality point clouds and generate new, diverse, faithful and visual-pleasing point clouds that exhibit evenly distributed points along the objects surface. It consists of a PointNet-based point cloud encoder, a style-aware decoder and a mapping network. Compared to state-of-the-art, our proposed framework is more lightweight, without complex compute-intensive modules, takes much less time in training while achieves state-of-the-art performance both in point cloud auto-encoding and generation. 

In the future, we would like to extend the work to more controllable design such as text guided shape generation. We also would like to extend the shape generation task in other forms such as mesh and implicit function for real world application usage. 


\bibliographystyle{IEEEtran}
\bibliography{tmm}
\end{document}